\pdfoutput=1
\documentclass[10pt, a4paper]{article}

\usepackage{lrec-coling2024}

\usepackage{latexsym}
\usepackage{multirow}
\usepackage{comment}
\usepackage[T1]{fontenc}
\usepackage[T5]{fontenc}
\usepackage[utf8]{inputenc}

\usepackage{microtype}
\usepackage{svg}
\title{Towards  Dog Bark Decoding: Leveraging Human Speech Processing  for Automated Bark Classification}
\name{Artem Abzaliev$^1$, Humberto Pérez Espinosa$^2$, Rada Mihalcea$^1$} 
\address{$^{1,3}$University of Michigan, $^2$National Institute of Astrophysics, Optics and Electronics (INAOE) \\
         abzaliev@umich.edu, humbertop@inaoep.mx, mihalcea@umich.edu.com\\}

\abstract{
Similar to humans, animals make extensive use of verbal and non-verbal forms of communication, including a large range of audio signals. In this paper, we address dog vocalizations and explore the use of self-supervised speech representation models pre-trained on human speech to address dog bark classification tasks that find parallels in human-centered tasks in speech recognition. We specifically address four tasks: dog recognition, breed identification, gender classification, and context grounding. We show that using speech embedding representations significantly improves over simpler classification baselines. Further, we also find that models pre-trained on large human speech acoustics can provide additional performance boosts on several tasks.  
 \\ \newline  }
\begin{document}
\maketitleabstract
\section{Introduction}

Until recently, ``what humans do'' has been considered the most widely accepted definition of intelligence \cite{tomasello2019becoming}, but a large body of recent work has demonstrated that there are numerous other forms of non-human intelligence \cite{bridle2022ways, Call2001DoAA, Biro2003CulturalIA}. While there are several new studies demonstrating plant intelligence \cite{wohlleben2016hidden, dosSantos2024PlantIH} %, calvo2020plants}, 
most of the research to date has focused on the intelligence of animals  \cite{de2016we,grandin2009animals}. Forms of animal intelligence range from memory \cite{matzel2010selective} and problem-solving \cite{seed2010problem}, %and knowledge generalization \cite{https://doi.org/10.1111/j.1748-7692.1994.tb00390.x}, 
all the way to the use of tools \cite{st2008revisiting} and communication \cite{seyfarth2003signalers, DazLpez2020WhenPM}. 

Like humans, animals use both verbal and non-verbal forms of communication, including audio signals such as calls, songs, or hisses; visual signals such as facial expressions, tail moves, or postural gestures; chemical cues; tactile cues; and bioluminescence. In general, the study of animal communication has been mainly addressed in fields such as biology, ecology, and anthropology, including, for instance, prairie dogs \cite{slobodchikoff2009prairie}, birds \cite{thorpe1961bird} or body movement in bees \cite{al2013honey}. Only recently we have started to see research that leverages advances in machine learning \cite{Bergler2019-cy, Jasim2022-gd, Maegawa2021-hz}.%, YIN2004343}.

% As discussed in \cite{YIN2004343}, specifically for dogs some researchers argued that barks are just attention-seeking vocalization, and do not include any context-specific information. However, as authors show this is not the case: dogs use different vocalization for different contexts, for instance more higher-pitch, modulated barks during the plays.

Focusing specifically on animal vocal communication, a recent study \cite{sperm_whales} highlighted three main questions to be answered to increase our understanding of how animals communicate: (1) What are the phonetic and perceptual units used by animals? (phonemes); (2) What are the composition rules used to combine those units? (morphology, ~syntax); and (3) Do those units carry meaning and, if so, how do we map the sound units to their meaning? (semantics, ~pragmatics). %There have been many attempts to answer those questions. For instance, back in 1923 \citet{old_bird_phonemes} tried to identify whether birds possess something similar to the phonemes, with more evidence and discussion thereafter by \cite{engesser2015experimental, bowling2015animal}.

In this work, we explore the third question and specifically attempt to understand the semantics of dog vocalizations. We use a state-of-the-art human speech representation learning model and show that such models can predict the context of a bark. %In future work, we plan to analyze the discrete units learned by the model and explore the correlation with the context.

This paper makes three main contributions. First, we introduce a dataset and a  set of tasks for dog bark classification. We draw parallels between human speech classification tasks and dog bark classification tasks, including dog recognition, breed recognition, gender identification, and context grounding.
Second, through several experiments, we show that we can leverage models developed for human speech processing to explore dog vocalizations and demonstrate that these can be used to significantly enhance performance on several dog bark classification tasks. Finally, through this work, we hope to open new opportunities for research in the area of animal communication, which can leverage the extensive expertise available in the NLP community.

% Please add the following required packages to your document preamble:
% \usepackage[normalem]{ulem}
% \useunder{\uline}{\ul}{}
\begin{table*}[]
\centering
\scalebox{0.8}{
\begin{tabular}{l|r r} \hline
\multicolumn{1}{c}{Context}                           & \# segments & Duration (sec) \\ \hline
Very aggressive barking at a stranger (L-S2)   & 2,843      & 2,778.66          \\ 
Normal barking at a stranger (L-S1)    & 2,772      & 2,512.92                  \\ 
Barking due to assault on the owner (L-A)   & 829       & 956.58              \\ 
Negative grunt (during the presence of a stranger) (GR-N)   & 637  & 746.60   \\ 
Negative squeal (during the presence of a stranger) (CH-N) & 298     & 546.72 \\ 
Sadness/anxiety barking (L-TA)  & 288       & 200.27                          \\ 
Positive squeal (during gameplay) (CH-P)  & 91    & 150.49                    \\ 
Barking during play (L-P)  & 76        & 51.21                                \\ 
Barking due to stimulation when walking (L-PA)  & 62        & 84.06           \\ 
Barking in fear at a stranger (L-S3)   & 54        & 45.08                    \\ 
Positive grunt (during gameplay) (GR-P)   & 51        & 79.56                 \\ 
Barking arrival of the owner at home (L-H)  & 24        & 26.20               \\ 
Barking that is neither playful nor strange (L-O)   & 9         & 9.50        \\ 
Non-dog sounds (voices, TV, cars, appliances, etc.) (S) & 8,755 & 14,304.05     \\ \hline
{\sc Total} & 16,789 & 22,491\\
\hline 
\end{tabular}
}
\caption{14 types of dog vocalizations together with the corresponding number of segments and duration.}
\vskip -0.2in
\label{tab:datalabels}
\end{table*}

\section{Related Work}

\paragraph{Animal Communication Datasets.} 
Compared to human languages, there are significantly fewer datasets available for animal communication. %One of the main reasons is that one cannot follow a typical human language dataset construction pipeline (collect raw data, hire annotators), and needs to either find creative ways to label the data or use self-supervised learning. This is a major shortcoming, slowing down progress in this field.
The largest library of animal vocalizations is the Macaulay Library at the Cornell Lab of Ornithology,\footnote{\url{https://www.macaulaylibrary.org}} which includes audio, photos, and videos of 2,674 species of amphibians, fish, mammals, and more, with the main focus of the library on birds.  Another large library of animal vocalizations is the Animal Sound Archive\footnote{\url{https://www.tierstimmenarchiv.de/webinterface/}}, which covers 1,800 bird species and 580 mammal species. %Again, the amount of dog vocalization is rather scarce, although it includes textual descriptions. 

There are also several datasets related to marine mammals. \citet{ness2013orchive} presented a  large dataset of over 20,000   recordings of Orca vocalizations. The Watkins Marine Mammal Sound Database\footnote{\url{https://cis.whoi.edu/science/B/whalesounds/index.cfm}} contains 15,000 annotated sound clips for more than 60 species of marine mammals.

Specifically for dog vocalizations, one of the most popular datasets was introduced by \citet{pongracz2005human}. It includes twelve Mudi dogs and consists of 244 recordings.  Another dataset is the UT3 database \citet{gutierrez2019classification}, with 74 dogs and ~6,000 individual audios. Neither of these datasets is publicly available. 

\paragraph{Computational Approaches to Animal Communication Analysis.}

Several studies have applied machine learning to animal communication, most of which used Convolutional Neural Networks (CNNs) to classify bird calls \citet{Maegawa2021-hz,Jasim2022-gd}, primate species \citet{Pellegrini2021-uw,Oikarinen2019-il},  multi-species classification of birds and frogs \citet{LeBien2020-xg}, or orca sounds \cite{Bergler2019-cy}. \citet{Ntalampiras2018-sh} used various methods to transfer the signal from music genre identification to bird species identification. %\citet{Maegawa2021-hz} and \citet{Jasim2022-gd} used CNN to classify bird calls, \citet{Pellegrini2021-uw} and \citet{Oikarinen2019-il} applied various CNNs to classify primate species. \citet{LeBien2020-xg} applied CNNs for multi-species classification of birds and frogs. \cite{Bergler2019-cy} used three separate neural networks to cluster orca sound types. \cite{Ntalampiras2018-sh} used various methods to transfer the signal from music genre identification to bird species identification. 

Specifically for dogs, there have been several studies studying dog vocalizations \cite{perez2019evaluation, perez2015automatic, gutierrez2019classification}. Our work is more closely related to \cite{YIN2004343}, where the contexts in which barking occurs are predicted, along with individual dog recognition. The results of the experiments of \citet{hantke2018my} confirmed that one can predict the context of the bark. \citet{MOLNAR2009198} also finds that the barks include information about the individual dog, as well as information about the context. However, no pre-trained models for dog vocalizations are currently available. To our knowledge, we are the first to use neural acoustic representations for tasks on dog vocalizations, and we are also the first to explore the use of human speech pre-training. 

% AA: I decided not to cite last two papers on transfer  learning that  you mentioned, since they are a bit weird.  One is using Alexnet in 2021-2022, and calls it transfer learning, and 
% another one is just too complicated - has like HMM, tons  of manual features etc, I  didn't really liked it. \cite{,Lu2021-tv}

\begin{comment}
\begin{table}[]
\centering
\scalebox{0.85}{
\begin{tabular}{l|r}
\hline
               & Counts         \\ \hline
Number dogs     & 74         \\ 
\hspace{0.2in} Gender: female & 49 \\
\hspace{0.2in} Gender: male & 25 \\
\hspace{0.2in} Breed: Chihuahua & 42 \\
\hspace{0.2in} Breed: French Poodle & 21 \\
\hspace{0.2in} Breed: Schnauzer & 11 \\

Number contexts & 14             \\ 
Vocalization segments      & 16,789  \\ 
\hspace{0.2in} Whines      & 389            \\ 
\hspace{0.2in} Growls      & 688            \\ 
\hspace{0.2in} Barks       & 6,957           \\ 
\hline
\end{tabular}
}
\caption{Overall dataset statistics.}
\label{tab:datastats}
\end{table}
\end{comment}

\section{Dataset}\label{sec:dataset}
We use a dataset consisting of recordings of 74 dogs, collected in Tepic (Mexico) and Puebla (Mexico), at the homes of the dogs' owners. A subset of this dataset was previously used by \citet{data_article}. %Mescalina Bark dataset
The dog vocalizations were recorded while being exposed to different stimuli (e.g., stranger, play, see Table \ref{tab:datalabels}). The recordings were conducted using a video camera Sony CX405 Handycam; in this work, we only use the audio recordings, obtained using the built-in microphone on the camera. The audio codec is A52 stereo with a sampling rate of 48,000 Hz and a bit rate of 256 kbps. The protocol for obtaining the dog vocalizations used in this study was designed and validated by experts in animal behavior from the Tlaxcala Center for Behavioral Biology in Mexico.

The dataset includes recordings of 48 female and 26 male dogs, mostly of three breeds: 42 Chihuahua, 21 French Poodles, and 11 Schnauzer. For mixed breeds, we first selected the breed mentioned. We focused on these breeds since they are among the most common domestic breeds in Mexican households. Given time and resource constraints during the data collection process, these breeds allowed for a broader choice of participants. The dog's average age is 35 months, ranging between 5 to 84 months old. 

\begin{figure}[t!]
    \centering
    \includegraphics[width=\columnwidth]{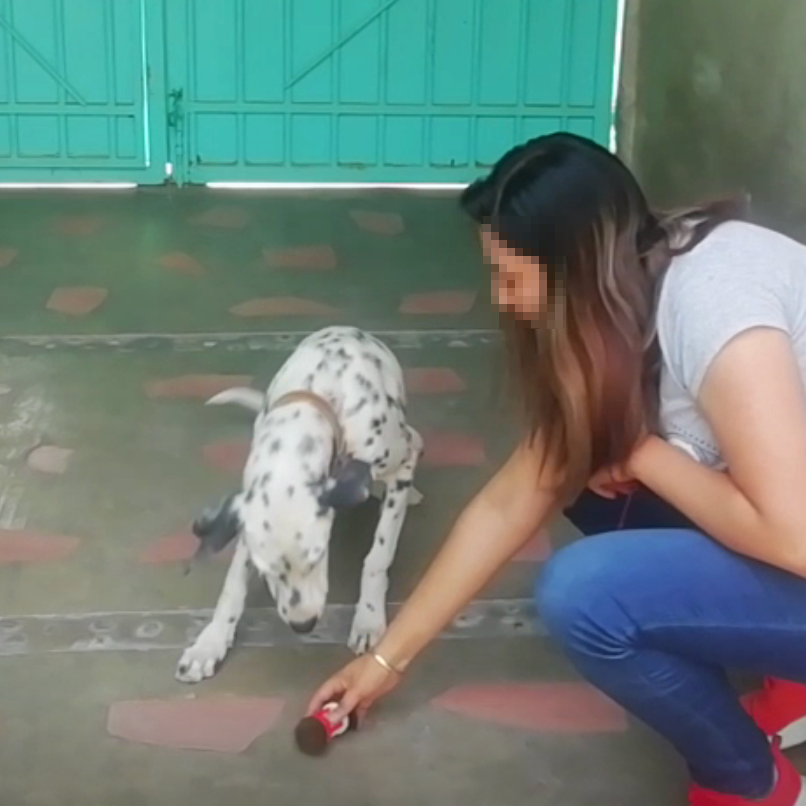}
    % \vspace{-0.4in} 
    \caption{Data collection for the stimulus ``playing with toy''; the owner stimulates the dog using the toys with which the dog normally plays.}
    \label{fig:data_collection}
\end{figure}

\paragraph{Stimuli.} Dog vocalizations were induced by exposing them to several stimuli, with the participation of the owner and/or an experimenter. To illustrate, the following represent examples of situations used during the data collection: the experimenter repeatedly rings the home doorbell and knocks the door hard; the experimenter simulates an attack on the owner; the owner speaks affectionately to the dog; the owner stimulates the dog using the objects or toys with which the dog normally plays; the owner performs the normal routine that precedes a walk; the owner ties the dog on a leash to a tree and walks out of sight (see Figure \ref{fig:data_collection} as an example); and others. The dogs are recorded while reacting to these stimuli, resulting in recordings lasting between 10 sec to 60 min. 

\paragraph{Data Processing and Annotation.} The recordings are automatically segmented into shorter segments ranging between 0.3 to 5 sec in length. The segmentation is performed using a threshold to separate between sound and silence or background noise; the threshold was identified using the short-time energy and spectral centroid aspects of the acoustic signal. Only the sound segments are used for the experiments. 
% These shorter segments range from 0.3 to 5 
Each of the resulting segments was manually annotated using the information associated with the stimulus. One of the fourteen contexts was assigned to each segment; if the audio did not have any dog-related sounds, it was assigned a Non-dog sound label. %Annotators used an open-source audio annotation software Audacity.% \cite{Audacity}. 

Table \ref{tab:datalabels} shows the fourteen labels used in the annotation, along with the corresponding statistics for the number of segments and total duration.

\section{Dog Bark Classification Tasks} 

Using acoustic representations of dog barks, we explore several fundamental tasks, including the recognition of individual dogs; the identification of the breed of a dog; the identification of a dog's gender; and the grounding of a dog bark to its context. These tasks have counterparts in human speech analysis, such as speaker identification or grounded language analysis.

\paragraph{Leveraging human speech for acoustic dog bark representations} To create acoustic representations of the dog vocalizations in the dataset, we fine-tune a pre-trained state-of-the-art self-supervised speech representation model. We use Wav2Vec2 \cite{baevski2020wav2vec}, which uses a self-supervised training objective to predict masked latent representations, pre-trained on the Librispeech corpus \cite{panayotov2015librispeech}. Wav2vec2 uses 960
hours of unlabeled human-speech data to learn how to represent audio signals as a sequence of discrete tokens. Some of those discrete tokens are masked, similar to the process used to train the BERT contextual embedding model \cite{devlin2018bert}. In Wav2Vec2, the learning of discrete units and unmasking are happening simultaneously.
%Wav2vec2 is a family of models pre-trained on a large corpus of unlabeled data on speech corpus Librispeech. Those models can be further fine-tuned on labeled data. For simplicity 

We use an open-source implementation of Wav2Vec2 from HuggingFace \cite{wolf2019huggingface}. We experiment with two model versions: (1) a model trained from scratch, using the dog vocalizations dataset in Section \ref{sec:dataset}; (2) a model pre-trained on 960 hours of unlabeled human speech data, and fine-tuned on dog vocalizations. 

\paragraph{Experimental Setup.} All the experiments use a ten-fold cross-validation setup. Specifically, for the tasks of breed identification, gender identification, and grounding, we use {\it grouped} ten-fold validation, with individual dogs being the group variable. That is, we leave 7-8 dogs as a test dataset and train on the remaining dogs' vocalizations, to control for any confounding information. For the dog recognition task, given the goal to recognize individual dogs, the model has to see each class (i.e., each dog) during the training, and thus all 74 individual dogs have to be present in both the training and test datasets. We note that this particular way of cross-validating might enable easier learning for the model and does not prevent shortcut learning, which is a common drawback for all author identification tasks. Therefore even Wav2Vec2 pretrained from scratch performs relatively well, and the performance boost is more pronounced than for other tasks. 
% \artem{can you please include 2-3 paragraphs of background on wav2vec, how it works;  why choose this representation over alternatives; data typically used for pre-training, what do we use}
% \artem{can we add here an explanation of (from scratch) and (pre-trained) -- described in the tables below}

\subsection{Dog Recognition}

We formulate this task as classifying a single audio segment as belonging to one of the 74 dogs in the dataset. According to \cite{molnar2006can} humans struggle to discriminate between individual dog barks, but machine learning methods, even unsupervised, can perform rather well \cite{YIN2004343}. This task is similar to identifying speakers, where many datasets \cite{Nagrani2017VoxCelebAL, Chung2018VoxCeleb2DS} and methods  \cite{huang2023masked, ding2020autospeech} already exist. 

Table \ref{tab:dog_recognition} shows the results,  where we apply the Wav2Vec2 model to dog identification. Our results are in line with the results from \cite{data_article,MOLNAR2009198}, and demonstrate that effectiveness of acoustic representations to discriminate between individual dogs. Further, we find that a model pre-trained on human speech significantly outperforms the model trained from scratch. 

Given the differences between human speech and animal vocalizations, we still need more work to understand how pre-training on human speech improves the performance on dog vocalization tasks. We believe that the pre-training on human speech enables the model to learn abstract vocalization structures, which in turn are beneficial for understanding animal vocalizations. This hypothesis is supported by previous studies showing that pre-training on seemingly unrelated tasks can be beneficial, for instance, pretraining on symbolic music data and applying it to natural language data provides significant performance improvements due to the ability of the neural networks used by the model to represent abstract syntactic structure \cite{papadimitriou2020learning}. Similarly, in computer vision, pre-training on ImageNet (object recognition) data is found to improve radiography processing \cite{jabbour2020deep}. 

\begin{table}[h]
\centering 
\scalebox{0.8}{
\begin{tabular}{lc}
%\multicolumn{2}{c}{Dog recognition (74 classes)} \\ 
\hline %\cline{2-5} 
Method & Accuracy \\ \hline
Majority & 5.03\%  \\ 
Wav2Vec2 (from scratch) & 23.74\% \\ 
Wav2Vec2 (pre-trained)   & \textbf{49.95\%} \\ \hline 
\end{tabular}
}
\caption{Accuracy for the dog recognition task.}
\vskip -0.2in
\label{tab:dog_recognition}
\end{table}

\begin{table*}
\centering 
\scalebox{0.8}{
\begin{tabular}{lcccc}
\hline 
        &         & \multicolumn{3}{c}{F-1 measure} \\ \cline{3-5}
Method &                 Acc. & Chihuahua & French Poodle & Schnauzer \\ \hline
Majority &                58.76\%   & 61.49\% &  6.59\% &  6.78\% \\ 
Wav2Vec2 (from scratch) & 60.18\%  & 74.42\% &   14.96\% & 5.79\% \\ 
Wav2Vec2 (pre-trained)   & \textbf{62.28\%} & \textbf{74.47\%} & \textbf{36.11\%} & \textbf{14.88\%}  \\ \hline 
\end{tabular}
}
\caption{Accuracy and F-1 measure for dog breed identification.}
%  More detailed results can be found in the Appendinx in Table \ref{tab:breedX}.
\vskip -0.1in
\label{tab:breed}
\end{table*}

\begin{table*}
\centering 
\scalebox{0.8}{
\begin{tabular}{lccccc}
\hline 
        &         & \multicolumn{4}{c}{F-1 measure} \\ \cline{3-6}
Method &                 Acc. & L-S2 & CH-N & GR-N & L-S1 \\ \hline
Majority &                56.37\% & 41.31\% & 0.00\% & 0.00\% & 30.39\%  \\ 
Wav2Vec2 (from scratch) & 58.45\% & 49.26\% &  21.26\% & 78.20\% & 48.64\% \\ 
Wav2Vec2 (pre-trained)   & {\bf 62.18\%} & \textbf{49.66\%} & {\bf 45.26\%} & {\bf 90.70\%} & {\bf 51.13\%} \\ \hline 
\end{tabular}
}
\caption{Accuracy and F-1 measure for context grounding.}
%  More detailed results can be found in the Appendix in Table \ref{tab:contexts}.
\vskip -0.1in
\label{tab:context}
\end{table*}

\subsection{Breed Identification}

In this task, we aim to predict the breed of a dog. Our dataset contains mostly three breeds: Chihuahua, French Poodle, and Schnauzer. %In the case of mixed-breed dogs, we select the first breed listed in the data description. %Note that this experiment is different from the experiment performed in \cite{data_article}: the authors build a separate model for each breed but predict an individual dog, while we are focusing on predicting the breed from the acoustic signal. 
We hypothesize that different breeds have different pitches so the acoustic model should be able to identify those differences, independent of the context. This experiment is related to previous work \cite{LeBien2020-xg,Oikarinen2019-il}. The task is similar to human accent recognition, where given audio files in a single language (i.e., English) the goal is to classify the accent of a speaker (e.g., USA vs. UK vs. India), with several approaches proposed in previous work \cite{ayranci2020speaker,honnavalli2021supervised,sun2002pitch}.

The results are shown in Table \ref{tab:breed}. Wav2Vec2 trained from scratch outperforms most baselines. As before, we obtain an additional significant boost in performance when pre-training on human speech data. The variation in individual breeds can be explained by the unbalanced number of observations per breed, with Chihuahua being the most common breed in our dataset (57\%), followed by French Poodle (28\%) and Schnauzer (15\%).   %As can be seen for the 

% \begin{table}
% \centering 
% \begin{tabular}{lc}
% \multicolumn{2}{c}{Breed identification (3 classes)} \\ \hline %\cline{2-5} 
% Method & Accuracy \\ \hline
% Majority & 58.76\%  \\ 
% Wav2Vec2 (from scratch) & 60.01\% \\ 
% Wav2Vec2 (pre-trained)   & \textbf{63.40\%} \\ \hline 
% \end{tabular}
% \caption{Accuracy results for the breed recognition task. The labels are Chihuahua, French poodle and Schanuzer. Only top 4 contexts are used.}
% \label{tab:breed_identification}
% \end{table}

% \begin{table*}[]
% \begin{tabular}{lcccc}
%                        & Good/bad bark            & Sentiment (with growl/squeals) & S1/S2           &  \\ \cline{1-4}
% Majority               & \textbf{77.84\%} (16.19) & \textbf{70.93\%} (10.67)       & 65.35\% (09.20) &  \\ 
% Wav2Vec2(scratch)      & 59.34\% (15.27)          & 68.11\% (13.92)                &               63.48 (06.18)  &  \\ 
% Wav2Vec2 (pre-trained) & 58.53\% (19.66)          & 64.70\% (16.21)                & \textbf{51.95\%} (12.69) &  \\ 
% \# examples per class  & 162/162                  & 410                            & 2843/2772       &  \\ \cline{1-4}
% \end{tabular}
% \end{table*}

\subsection{Gender Identification}

The goal of this task is to probe whether it is possible to predict the gender of a dog from vocalizations. This is a task analogous to the prediction of   demographics (e.g., age, gender) from language or speech, with  many previous studies  conducted on this topic %\cite{qawaqneh2017deep, ghahremani2018end, 
\cite{qawaqneh2017deep, saraf2023zero, gupta2022estimation, welch-etal-2020-compositional}. 

\begin{table}[h]
\centering 
\scalebox{0.8}{
\begin{tabular}{lccc}
\hline 
        &         & \multicolumn{2}{c}{F-1 measure} \\ \cline{3-4}
Method &                 Acc. & Female & Male \\ \hline
Majority &                68.70\%   & 74.73\% &  7.76\%  \\ 
Wav2Vec2 (from scratch) & {\bf 70.07\%}  & 76.54\% &   {\bf 19.29\%} \\ 
Wav2Vec2 (pre-trained)   & 68.90\% & \textbf{79.31\%} & 7.10\% \\ \hline 
\end{tabular}
}
\caption{Accuracy and F-1 measure for dog gender identification.} 
%  More detailed results can be found in the Appendix in Table \ref{tab:genderX}.
\vskip -0.1in
\label{tab:gender}
\end{table}

Table \ref{tab:gender} shows the results. The Wav2Vec2 model trained from scratch performs better than the baseline model, with no further improvements obtained with Wav2Vec2 pre-trained on human speech. Interestingly, we do see an improvement brought by human speech pre-training on the female class, for which we have significantly more data in our dataset (67.95\% female vs 32.04\% male by total duration). We found that gender identification is the most difficult task among all the tasks we propose. We hypothesize that the model trained from scratch focuses on learning acoustic features, while the pre-trained wav2vec attempts to learn shortcuts and overfits quickly. We noticed that it often predicts just the majority class (female) so that F1 increases for female and decreases for male, while the overall accuracy is almost the same as for the majority baseline.%We interpret this as a result of an unbalanced dataset: we have 48 Female species and 26 Male species. 

% \artem{48 + 26 != 74? (no. of dogs stated at the beginning of the Dataset section}

% TODO -- could also merge this subsection with the one before and rename it Demographic Identification, and separately talk about breed and gender

% \subsection{Detection of Emotion}

% -- describe the task formulation, use Experiment 1 split from sec 6 in \cite{data_article}, give intuition as to why it should work

% -- make analogy with human emotion classification, give citation (eg, AffectiveText task from Semeval, work on emotion in speech, etc.)

% -- show table of results; include results per class negative / positive

% -- can include also results with the Experiment 2 and Experiment 3 splits in sec 6?

% Please add the following required packages to your document preamble:
% \usepackage{multirow}

\subsection{Grounding}

In this task, we predict the context of the bark; i.e., we  determine the association between a dog vocalization and its surrounding. Because of the highly skewed label distribution (see Table \ref{tab:datalabels}), we focus on the contexts for which more examples are available: very aggressive barking at a stranger (L-S2); normal barking at a stranger (L-S1); negative squeal (in the presence of a stranger) (CH-N); negative grunt (in the presence of a stranger) (GR-N). We do not include barking due to assault on the owner (L-A) because in early experiments we found that the model cannot distinguish it from the very aggressive barking at a stranger (L-S2).
 
 Human language grounding is the mapping of language symbols such as words to their corresponding objects in the real world. There have been several works showing that animals ground their vocalizations as well. For instance, the vocalizations of prairie dogs are grounded and used to transmit   the characteristics of the  predators (e.g., color or size) \cite{slobodchikoff2009prairie}. Other work  has also demonstrated that it is possible to predict call types for marmoset monkeys  \cite{Oikarinen2019-il} also shows. We hypothesize that dog vocalizations are related to their context, and therefore can be grounded. 

Table \ref{tab:context} shows the results. Similar to the previous experiments, both Wav2Vec2 models outperform the majority baseline, with the Wav2Vec2 pre-trained on human speech leading to the most accurate results.

% Language grounding \artem{add 1-2 sentences about human language grounding}.
% -- talk about how this is analogous to human language description of context, make connection to grounding, give citations

\section{Conclusion}
% is related to the past work + how do we answer about the meaning. 

In this paper, we explored the use of pre-trained self-supervised speech representation models to address dog barking classification tasks. We specifically addressed four tasks that find parallels in human-centered speech recognition tasks: dog recognition, breed recognition, gender identification, and context grounding. We showed that acoustic representation models using Wav2Vec2 can significantly improve over simpler classification baselines. Additionally, we found that a model pre-trained on human speech can further improve the results. We hope  our work will encourage others in the NLP community to start addressing the many research opportunities that exist in the area of animal communication. The dataset used in this work, along with the baselines that we introduced, are publicly available by request from \texttt{humbertop@ccc.inaoep.mx}.%The dataset described in this paper is publicly available for research purposes. 

\section{Limitations}

In this work, we focused on only one species, domestic dogs, and only three breeds. More species are required to understand how modern computational methods can be used for studying animal vocalization. In the future, we are planning to extend our work to birds and marine mammals, since those species have a large amount of data available.

We also focused on only one  neural network architecture, Wav2Vec2. While it is a popular architecture for human speech processing, other architectures might be more suitable for studying animal vocalizations. Also, we used supervised learning in this work, since the dataset was manually annotated. The majority of the datasets are not annotated and thus would require semi-supervised or unsupervised learning, which is more challenging.
\bibliography{custom}
\bibliographystyle{acl_natbib}

\end{document}